\pdfoutput=1
\documentclass[11pt]{article}

\usepackage[]{acl}

\usepackage{bm}
\usepackage{times}
\usepackage{latexsym}
\usepackage{graphicx}
\usepackage{multirow}
\usepackage{color}
\usepackage{makecell}
\usepackage{booktabs}

\usepackage[T1]{fontenc}
\usepackage[utf8]{inputenc}

\usepackage{microtype}

\title{Prompt-based Connective Prediction Method for Fine-grained Implicit Discourse Relation Recognition}

\author{
    Hao Zhou$^{1}$, Man Lan$^{1*}$, Yuanbin Wu$^{1}$, Yuefeng Chen$^2$ and Meirong Ma$^2$ \\ 
    $^1$School of Computer Science and Technology, \\ East China Normal University, Shanghai, China \\
    $^2$Shanghai Transsion Co., Ltd., Shanghai, China\\
    \texttt{hzhou@stu.ecnu.edu.cn,\{mlan,ybwu\}@cs.ecnu.edu.cn} \\ 
    \texttt{\{yuefeng.chen,meirong.ma\}@transsion.com}
}

\begin{document}
\maketitle
\begin{abstract}
Due to the absence of connectives, implicit discourse relation recognition (IDRR) is still a challenging and crucial task in discourse analysis. Most of the current work adopted multi-task learning to aid IDRR through explicit discourse relation recognition (EDRR) or utilized dependencies between discourse relation labels to constrain model predictions. But these methods still performed poorly on fine-grained IDRR and even utterly misidentified on most of the few-shot discourse relation classes. To address these problems, we propose a novel \textbf{P}rompt-based \textbf{C}onnective \textbf{P}rediction (\textbf{PCP}) method for IDRR. Our method instructs large-scale pre-trained models to use knowledge relevant to discourse relation and utilizes the strong correlation between connectives and discourse relation to help the model recognize implicit discourse relations. Experimental results show that our method surpasses the current state-of-the-art model and achieves significant improvements on those fine-grained few-shot discourse relation. Moreover, our approach is able to be transferred to EDRR and obtain acceptable results. Our code is released in \url{https://github.com/zh-i9/PCP-for-IDRR}.
\end{abstract}

\section{Introduction}
Discourse relations recognition (DRR) aims to identify the semantic relation between two discourse units (e.g., sentences or clauses, they are denoted as \emph{Arg1} and \emph{Arg2} respectively). DRR is essential to many natural language processing (NLP) downstream tasks involving more context, such as document-level machine translation \citep{xiong2019modeling} and machine reading comprehension \citep{mihaylov2019discourse}, and text summarization \citep{xu2019discourse}.

\begin{figure}[tp]
    \centering
    \includegraphics[scale=0.55]{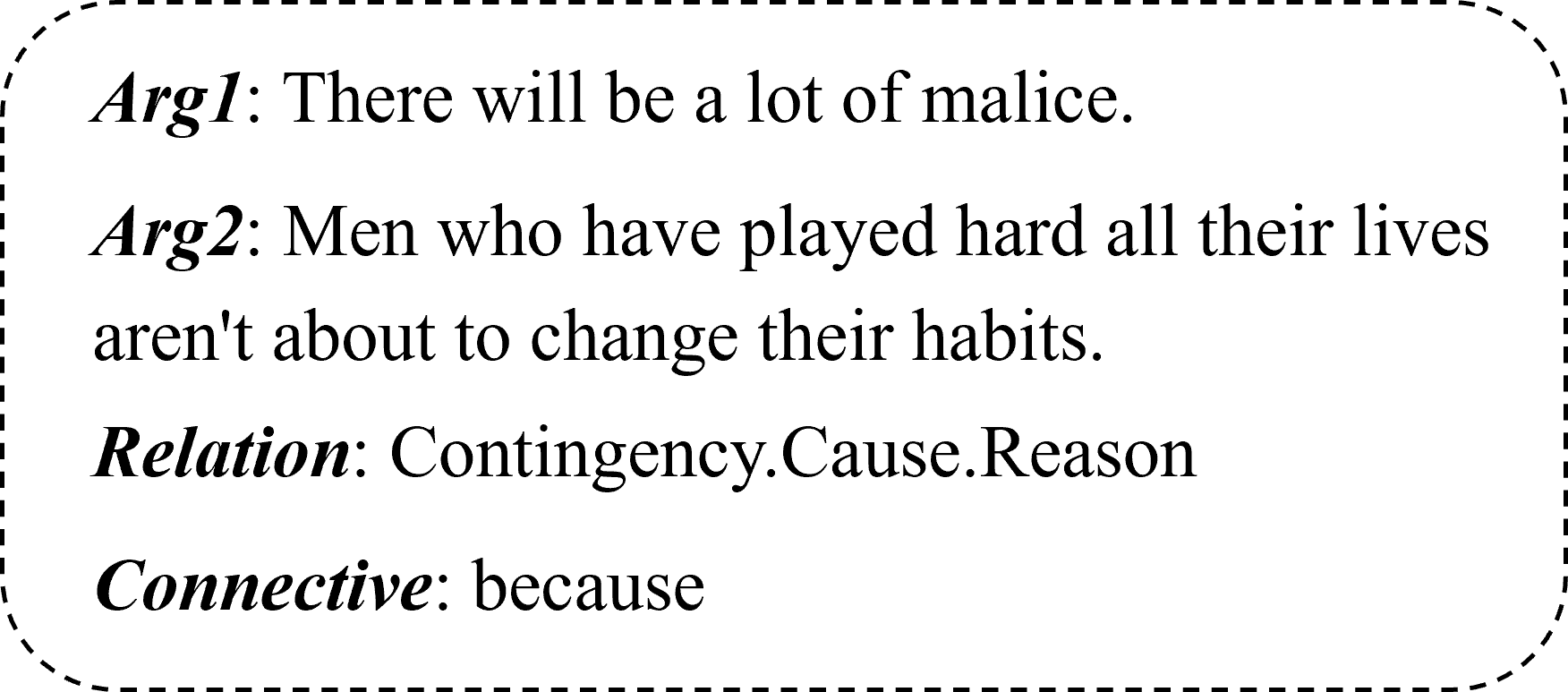}
    \caption{An IDRR instance in PDTB 2.0. For IDRR, the \emph{Connective} are not present in the original discourse context, but are assigned by the annotator according to semantic relations between \emph{Arg1} and \emph{Arg2}.}
    \label{fig:data_sample}
\end{figure}

Compared with explicit discourse relation recognition (EDRR), due to the absence of explicit connectives, implicit discourse relation recognition (IDRR) is more challenging and attractive to researchers. Figure \ref{fig:data_sample} shows an IDRR sample in the Penn Discourse Treebank 2.0 (PDTB 2.0) \citep{prasad-etal-2008-penn}.

Traditional machine learning based methods primarily relied on human-designed and shallow linguistic features \citep{lin2009recognizing,zhou2010effects}. With the rapid development of deep learning, a large number of methods prefer to utilize deep neural networks to recognize discourse relation between two arguments \citep{liu2016implicit,liu2016recognizing,dai2018improving}. Recently, pre-trained language models (PLMs), such as BERT \citep{devlin2018bert} and RoBERTa \citep{liu2019roberta}, play a dominative role in many NLP tasks through contextual representation learning. Leveraging the success of PLMs, extensive research work design a wide variety of post-processing neural networks \citep{liu2020importance,wu2021label} to extract the semantic information by fine-tuning them on the task-specific dataset.

On the other side, previous work showed the accuracy of explicit discourse relation recognition achieved more than 93\% using the discourse connective alone \citep{pitler-nenkova-2009-using}, which reveals strong correlations between connectives and discourse relations. Therefore, more and more studies focus on implicit discourse relation recognition through connective prediction. 
\citet{kishimoto-etal-2020-adapting} applied an auxiliary pre-training task, which utilized the representation of the [CLS] token to predict explicit connectives, to BERT using domain text containing explicit connectives. \textbf{Since} the idea of their method is contradictory with the original next sentence prediction (NSP) task of BERT and does not make use of the knowledge that BERT had learned through the masked language model (MLM) task. The experimental results showed that their method does not perform well on coarse-grained and fine-grained discourse relations. 
\citet{kurfali2021let} came up with a pipeline approach, where a connective selected from candidates and two implicit arguments were first integrated as “\emph{Arg1} \emph{Connective} \emph{Arg2}” sequence. Then this integrated sequence was input into an explicit discourse relation classifier trained in explicit data to identify implicit discourse relations in a distant supervision manner. \textbf{Since} there are a lot of connective candidates (about 100 in PDTB 2.0), and it requires a large number of model inferences for each sample. Obviously, the simple sequence does not make full use of the semantic knowledge embedded in the PLMs. 
Therefore, how to make better connective prediction becomes a crucial step for implicit discourse relation recognition.

Inspired by \citet{schick2020small}, we propose a novel \textbf{P}rompt-based \textbf{C}onnective \textbf{P}rediction (\textbf{PCP}) method for implicit discourse relation recognition. We exploit the advantage of prompt learning \citep{liu2021pre} to bridge the gap between connective prediction in the pre-training and fine-tuning stage and make better use of the knowledge of PLMs. 
Specifically, we \textbf{first} manually design templates matching natural language patterns, which elicit large-scale pre-trained language models for connective prediction using specific knowledge (in Section \ref{sec:prompt_construction}). 
\textbf{Secondly}, we select the less ambiguous connectives corresponding to each discourse relation based on the frequency of the connectives in the dataset to implement answer mapping (mapping predicted connectives to discourse relaiton labels) (in Section \ref{sec:answer_search}). 
Experimental results prove that the simple but proper template can outperform the current state-of-the-art (SOTA) model based on fine-tuning and achieve zero breakthroughs on few-shot fine-grained discourse relation. Furthermore, our approach based on connective frequency effectively avoids the trouble of manually selecting answer words and takes full advantage of the relevance between discourse relations and connectives. 

Our contributions are summarized as follows:
\begin{itemize}
  \item We propose a Prompt-based Connective Prediction (PCP) method for implicit discourse relation recognition, which achieves SOTA performance on the PDTB 2.0 dataset and CoNLL-2016 Shared Task as well.
  \item The method we proposed breaks the bottleneck of previous work on the few-shot fine-grained discourse relation of PDTB 2.0.
  \item Our approach can be easily transferred from IDRR to EDRR, and we have experimentally demonstrated that our approach still performs well for EDRR.
\end{itemize}
% 贡献总结：
% 1.提出了基于模板的连接词预测方法，充分地利用了连接词和语篇关系的联系来做IDRR，有效的提升了IDRR任务的性能，在两个数据集上实现了sota的效果
% 2.在小样本类别上实现了零的突破
% 3.可以将我们的方法迁移到显示语篇关系上 仍旧有不错的性能

\section{Related Work}

\subsection{Implicit Discourse Relation Recognition}
Previous methods based on machine learning primarily relied on manually-designed linguistic features \citep{lin2009recognizing,zhou2010effects}. With the rapid development of deep learning, most of methods prefer to utilize deep neural networks, such as RNN \citep{liu2016implicit}, CNN \citep{varia-etal-2019-discourse}, or LSTM \citep{liu2016recognizing,lan2017multi,dai2018improving}, to recognize discourse relation between two arguments. 

Along with the booming development of pre-trained language models, most work designs various post-processing neural networks for information interaction to improve overall performance by fine-tuning their models \citep{van2019employing,wu2020hierarchical}. For example, \citet{liu2020importance} combines different levels of representation learning modules to address implicit discourse relation recognition. \citet{wu2021label} designs a label attentive encoder to learn the global representation of an input instance and its level-specific context and employs a label sequence decoder to output the predicted labels in a top-down manner.

Recently, several methods have emerged for implicit discourse relation recognition through connective prediction \citep{kishimoto-etal-2020-adapting,kurfali2021let}. They use efficient pre-trained language models and additional explicit data, but their performance is less than ideal. We think that their methods are contradictory to the original pre-training task, and there is a distribution difference between the explicit and implicit data.

\subsection{Prompt-based Models}
Prompt-based methods have received considerable attention with the emergence of GPT-3 \citep{brown2020language}. A series of research work \citep{schick2020small,chen2022adaprompt} have demonstrated that prompt learning can effectively stimulate knowledge from PLMs compared with conventional fine-tuning. Therefore, many prompt-based models have been proposed and they achieved outstanding performance in widespread NLP tasks, such as text classification \citep{schick2020exploiting,hu2021knowledgeable}, text matching \citep{jiang2022promptbert}, named entity recognition \citep{cui2021template,chen2021lightner}, and relation extraction \citep{Han2021PTRPT,chen2022knowprompt}, particularly on few-shot and zero-shot settings. The main idea of their approaches is to convert the downstream task into a cloze-style task that is closer to the pre-training step of PLMs with prompts.

As far as we know, we are the first to apply prompt learning \citep{schick2020small} for implicit and explicit discourse relation recognition task. 

\section{The Prompt-based Connective Prediction Approach}
\begin{figure*}[htbp]
    \centering
    \includegraphics[scale=0.6]{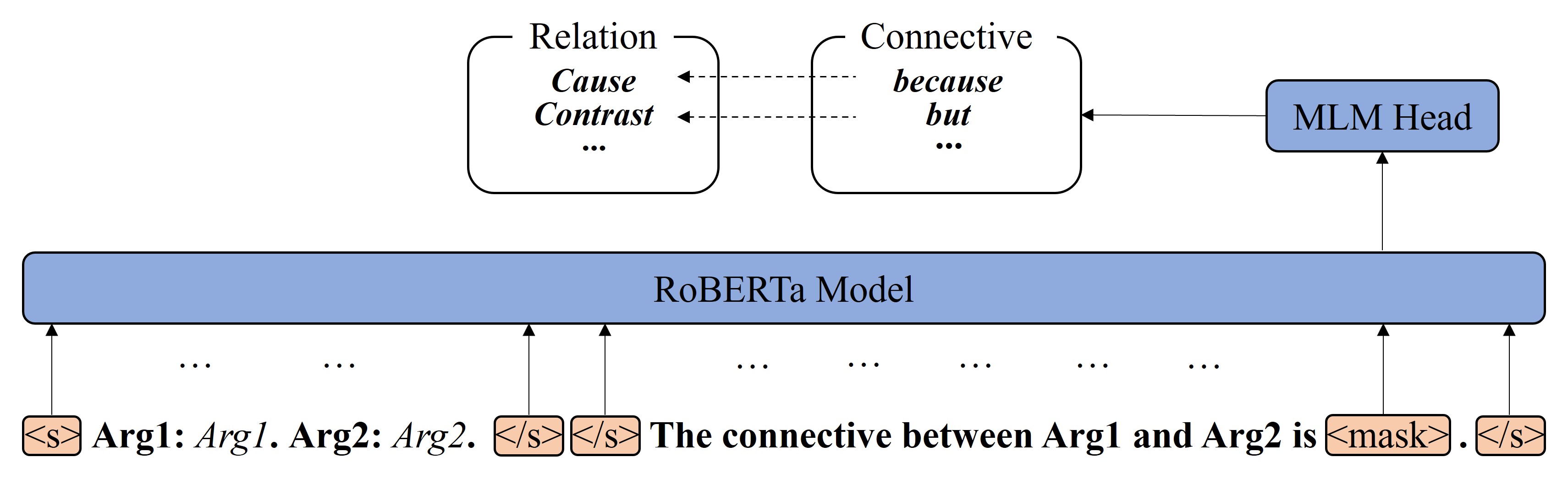}
    \caption{The model architecture of our Prompt-based Connective Prediction method.}
    \label{fig:model}
\end{figure*}

In this section, we will introduce our prompt-based connective prediction (PCP) approach to recognize implicit discourse relations in detail. Figure \ref{fig:model} shows the overall architecture of our model. 

% 『h』当前位置。将图形放置在正文文本中给出该图形环境的地方。如果本页所剩的页面不够，这一参数将不起作用。
% 『t』顶部。将图形放置在页面的顶部。
% 『b』底部。将图形放置在页面的底部。
% 『p』浮动页。将图形放置在一只允许有浮动对象的页面上。

\subsection{Connective Prediction with Prompt}
For convenience and clarity, we discuss our method with a simple template “\textit{Arg1} <mask> \textit{Arg2}.”. In this template, \emph{Arg1} and \emph{Arg2} correspond to two arguments in the original texts respectively (as shown in Figure \ref{fig:data_sample}), and the symbol <mask> represents the masked token in place of their connective. Given a pair of arguments $x_{arg1}$ and $x_{arg2}$, we transfer them to $x_{prompt}$ with the template:
\begin{equation}
    x_{prompt} = \mathbf{T}(x_{arg1},x_{arg2})
\end{equation}
where $\mathbf{T}$ represents template function. Then we feed $x_{prompt}$ to the RoBERTa \citep{liu2019roberta} model to obtain the representation of <mask> token $\bm{h}_{mask}$:
\begin{equation}
    \bm{h}_{mask} = \mathbf{RoBERTa}(x_{prompt})
\end{equation}
Same as MLM task in pre-training step, we input $\bm{h}_{mask}$ into MLMHead model, which predicts scores of the language modeling head (scores for each vocabulary token before $SoftMax$), and acquire the output $\bm{e}_{mask}$:
\begin{equation}
    \bm{e}_{mask} = \mathbf{MLMHead}(\bm{h}_{mask})
\end{equation}
We select the token with the highest score in $\bm{e}_{mask}$ as the connective predicted by the model. During the training, we use cross-entropy to calculate the loss between the model prediction and the golden connective:
\begin{equation}
    Loss = CE(target,SoftMax(\bm{e}_{mask}))
\end{equation}
where $CE$ is cross-entropy loss function and $target$ is golden connective (mapped from golden discourse relation label) index in the vocabulary. Finally, we are mapping the predicted connective (e.g., \emph{because}) to the corresponding discourse relation label (e.g., \emph{Cause}).

Then, we will detail the two above-mentioned crucial parts in our methods, \textbf{prompt construction} (Section \ref{sec:prompt_construction}) and \textbf{answer search} (Section \ref{sec:answer_search}).

\subsection{Prompt Construction}
\label{sec:prompt_construction}

\begin{table*}[htbp]
\centering
\resizebox{\linewidth}{!}{
\begin{tabular}{lc}
\toprule
\textbf{Template} & \textbf{PDTB Top-level Acc.}   \\
\midrule
\textit{Arg1} <mask> \textit{Arg2}.                                                                & 69.23   \\
\textit{Arg1}. That's <mask> \textit{Arg2}.                                                        & 67.37   \\
\midrule
Arg1: \textit{Arg1}. Arg2: \textit{Arg2}. The connective between Arg1 and Arg2 is <mask>.          & 71.09   \\
Arg1: \textit{Arg1}. Arg2: \textit{Arg2}. The conjunction between Arg1 and Arg2 is <mask>.         & 71.43   \\
Arg1: \textit{Arg1}. Arg2: \textit{Arg2}.</s></s>The connective between Arg1 and Arg2 is <mask>.   & 71.68   \\
Arg1: \textit{Arg1}. Arg2: \textit{Arg2}.</s></s>The conjunction between Arg1 and Arg2 is <mask>.  & 72.27   \\
\bottomrule                            
\end{tabular}}
\caption{Searching templates on \textit{RoBERTa-large}. The symbol </s> and <mask> represents segment and mask token in RoBERTa tokenizer.}
\label{table:template}
\end{table*}

For prompt-based methods, one key challenge is to find an appropriate template matching the target task. In this work, we transfer implicit discourse relation recognition to a connective prediction task and utilize manual search to identify suitable templates for the connective prediction task. Moreover, we use the accuracy of the top-level senses of the PDTB 2.0 development set as the main metric to evaluate different templates. Table \ref{table:template} shows our designed templates based on the frequent position of connectives and natural language instructions. 

In PDTB 2.0 dataset, for all implicit discourse relation samples, their connective position is between \textit{Arg1} and \textit{Arg2}. In other words, they all satisfy the “\textit{Arg1} connective \textit{Arg2}” sequence order. Therefore, we designed the first two templates in Table \ref{table:template} based on the position of connectives. Perhaps because there are some connectives that may not be appropriate for the sentence form “That's \textit{connective} ...” (e.g., \textit{but}), the accuracy value of the second template is lower than the first.

Inspired by \citep{schick2020small}, we design templates conforming to natural language patterns to instruct PLMs to output the content we want to obtain. In this work, our target is connective prediction using PLMs and then mapping connective to the final discourse relation label. Therefore, we use the template “The connective (or conjunction) between Arg1 and Arg2 is <mask>.” to direct PLMs for connective prediction. Meanwhile, we also attempt to place segment token </s> between two arguments and instruction sentences, which is to allow the model to take the coherence of overall inputs into account. We can see that the last template achieves the best performance on the accuracy value, and our final experimental results are also based on this template (as shown in Figure \ref{fig:model}).

\subsection{Answer Search}
\label{sec:answer_search}

For other prompt-based methods, like \citet{schick2020exploiting}, they regarded “terrible”, “bad”, “okay”, “good”, and “great” as the answer word to the rating that a customer gave to a restaurant on a 1- to 5-star scale based on their review’s text. They artificially selected answer words that appear to fit the golden label, but these words may not fit well with most of the samples in the dataset. 

For implicit discourse relation data, each sample has been annotated with the connective appropriate to this sample on PDTB 2.0 and CoNLL16 dataset (detailed in Section \ref{sec:dataset}). As a result, we select the \textbf{most frequent} and \textbf{less ambiguous} connectives as the answer words of the corresponding discourse relations based on the frequency of connectives in each discourse relation and the repetition rate of some connectives in different discourse relations. At the same time, to unify the model's input and output and facilitate the mapping between answer words and discourse relation labels, we select those connectives that are tokenized as a single token by the RoBERTa tokenizer.

\begin{table}[tbp]
\centering
\resizebox{\linewidth}{!}{
\begin{tabular}{c|c|c}
\toprule
\textbf{Top-level} & \textbf{Second-level} & \textbf{Answer Set}  \\
\hline
\multirow{2}{*}{Comparison}  & Concession         & although, nevertheless \\ \cline{2-3}
                             & Contrast           & but, however           \\ 
\hline                    
\multirow{2}{*}{Contingency} & Cause              & \begin{tabular}[c]{@{}c@{}}because, as, so,\\ consequently, thus\end{tabular}           \\ \cline{2-3}
                             & Pragmatic cause    & since                 \\
\hline                            
\multirow{5}{*}{Expansion}   & Alternative        & instead, rather, or     \\ \cline{2-3}
                             & Conjunction        & and, also, furthermore  \\ \cline{2-3}
                             & Instantiation      & instance, example      \\ \cline{2-3}
                             & List               & first                 \\ \cline{2-3}
                             & Restatement        & indeed, specifically   \\
\hline
\multirow{2}{*}{Temporal}    & Asynchronous       & \begin{tabular}[c]{@{}c@{}}then, subsequently,\\ previously, earlier, after\end{tabular}   \\ \cline{2-3}
                             & Synchrony          & meanwhile             \\
\bottomrule                             
 \end{tabular}}
\caption{Mapping between implicit discourse relation labels and connectives on PDTB 2.0 dataset, which has four top-level and 11 second-level senses. The answer set of top-level senses is a union set of second-level.}
\label{table:mapping_pdtb}
\end{table}

Table \ref{table:mapping_pdtb} and \ref{table:mapping_conll} show the final answer sets we selected based on the above conditions on PDTB 2.0 and CoNLL16 dataset. It is worth mentioning that some second-level senses have subtypes (also be called the third-level senses). For examples, the second-level sense \emph{Cause} has two subtypes \emph{Reason} and \emph{Result}. We select the connectives for each subtype and merge them into the second-level senses. Moreover, for those samples whose connectives are not in the corresponding answer set, we select the first connective of the corresponding answer set as the golden connective of this sample.

\begin{table}[tbp]
\centering
\resizebox{\linewidth}{!}{
\begin{tabular}{c|c}
\toprule
\multicolumn{1}{c|}{\textbf{Cross-level Senses}}                 & \textbf{Answer Set}               \\
\hline
Comp.Concession                    & \begin{tabular}[c]{@{}c@{}}although,\\nevertheless\end{tabular}   \\
\hline
Comp.Contrast                      & but, however              \\
\hline
Cont.Cause.Reason                  & because, as               \\
\hline
Cont.Cause.Result                  & \begin{tabular}[c]{@{}c@{}}so, thus,\\consequently\end{tabular}     \\
\hline
Cont.Condition                     & if                       \\
\hline
Exp.Alternative                    & unless, or                \\
\hline
Exp.Alternative.Chosen alternative & instead                  \\
\hline
Exp.Conjunction                    & \begin{tabular}[c]{@{}c@{}}and, also, \\ furthermore\end{tabular}     \\
\hline
Exp.Exception                      & rather                   \\
\hline
Exp.Instantiation                  & instance, example         \\
\hline
Exp.Restatement                    & specifically             \\
\hline
Temp.Asynchronous.Precedence       & then, subsequently        \\
\hline
Temp.Asynchronous.Succession       & \begin{tabular}[c]{@{}c@{}}previously, \\ earlier, after\end{tabular} \\
\hline
Temp.Synchrony                     & meanwhile                \\
\hline
EntRel                             & none                     \\
\bottomrule
\end{tabular}}
\caption{Mapping between implicit discourse relation labels and connectives on CoNLL16 dataset which has 15 cross-level senses. The \textit{EntRel} sense represents that there is no discourse relation between two arguments, but there is a relation between the entities in them.}
\label{table:mapping_conll}
\end{table}

\section{Experiments}

\subsection{Dataset}
\label{sec:dataset}

\paragraph{The Penn Discourse Treebank 2.0 (PDTB 2.0)}PDTB 2.0 is a large scale corpus annotated with information related to discourse relation, containing 2,312 Wall Street Journal (WSJ) articles \citep{prasad-etal-2008-penn}. PDTB 2.0 has three senses levels (i.e., classes, types, and sub-types). We follow \citet{Ji2015OneVI} to take the sections 2-20 as the training set, 0-1 as the development set, and 21-22 as the testing set. Meanwhile, we evaluate our model on the four top-level implicit classes and the 11 major second-level implicit types \citep{varia-etal-2019-discourse,liu2020importance,wu2021label}. Table \ref{table:pdtb_top} and \ref{table:pdtb_second} show the detailed statistics of the top-level and second-level senses respectively\footnote{We found that there are some data samples with two senses. In our data statistics and experiments process, we uniformly considered the first sense of these samples as their golden label for avoiding ambiguity. The same operation was performed on CoNLL16 dataset.}.

\paragraph{The CoNLL 2016 Shared Task (CoNLL16)}The CoNLL 2016 shared task \citep{xue2016conll} provides more abundant annotation than PDTB for shadow discourse parsing. The PDTB section 23 and Wikinews texts following the PDTB annotation guidelines were organized as the test sets. CoNLL16 merges several labels of PDTB. For example, \emph{Contingency.Pragmatic cause} is merged into \emph{Contingency.Cause.Reason} to remove the former type with very few samples. Finally, there is a flat list of 15 sense classes to be classified, detailed senses as shown in Table \ref{table:mapping_conll} the first column.

\begin{table}[tbp]
\centering
\begin{tabular}{crrr}
\toprule
\textbf{Top-level Senses} & \textbf{Train} & \textbf{Dev.} & \textbf{Test} \\ 
\midrule
Comparison (Comp.)               & 1894           & 191           & 146           \\
Contingency (Cont.)              & 3281           & 287           & 276           \\
Expansion (Exp.)                 & 6792           & 651           & 556           \\ 
Temporal (Temp.)                 & 665            & 54            & 68            \\ 
\midrule
Total                     & 12632          & 1183          & 1046          \\
\bottomrule
\end{tabular}
\caption{The implicit data statistics of top-level senses in PDTB 2.0.}
\label{table:pdtb_top}
\end{table}

\begin{table}[tbp]
\centering
\begin{tabular}{crrr}
\toprule
\multicolumn{1}{c}{\textbf{Second-level Senses}} & \textbf{Train}          & \textbf{Dev.}          & \textbf{Test} \\ 
\midrule
Comp.Concession              & 180            & 15            & 17            \\
Comp.Contrast                & 1566           & 166           & 128           \\
Cont.Cause                   & 3227           & 281           & 269           \\ 
Cont.Pragmatic cause         & 51             & 6             & 7             \\
Exp.Alternative              & 146            & 10            & 9             \\
Exp.Conjunction              & 2805           & 258           & 200           \\
Exp.Instantiation            & 1061           & 106           & 118           \\
Exp.List                     & 330            & 9             & 12            \\
Exp.Restatement              & 2376           & 260           & 211           \\
Temp.Asynchronous            & 517            & 46            & 54            \\
Temp.Synchrony               & 147            & 8             & 14            \\ 
\midrule
\multicolumn{1}{c}{Total}                        & 12406          & 1165          & 1039          \\
\bottomrule
\end{tabular}
\caption{The implicit data statistics of second-level senses in PDTB 2.0.}
\label{table:pdtb_second}
\end{table}

\begin{table*}[tbp]
\centering
\resizebox{\textwidth}{!}{
\begin{tabular}{lcccccc}
\toprule
\multicolumn{1}{c}{\multirow{2}{*}{\textbf{Model}}} & 
\multicolumn{2}{c}{\textbf{PDTB-Top}} & \multicolumn{2}{c}{\textbf{PDTB-Second}} & \textbf{CoNLL-Test} & \textbf{CoNLL-Blind} \\
  & Macro-F1    & Acc.   & Macro-F1    & Acc.   & Acc.   & Acc.   \\
\midrule
NNMA \citep{liu2016recognizing}               & 46.29          & 57.57          & -              & -              & -              & -              \\
ESDP \citep{wang2016two}                      & -              & -              & -              & -              & \textbf{40.91} & 34.20          \\
MANN \citep{lan2017multi}                     & 47.80          & 57.39          & -              & -              & 39.40          & \textbf{40.12} \\
PDRR \citep{dai2018improving}                 & 48.82          & 57.44          & -              & -              & -              & -              \\
RWP-CNN \citep{varia-etal-2019-discourse}     & \textbf{50.20} & \textbf{59.13} & -              & -              & 39.39          & 39.36          \\
\midrule
DER \citep{bai2018deep}                       & 51.06          & -              & -              & 48.22          & -              & -              \\
ELMo-C\&E \citep{dai2019regularization}       & 52.89          & 59.66          & -              & 48.23          & -              & -              \\
MTL-MLoss \citep{van2019employing}            & 53.00          & -              & -              & 49.95          & -              & -              \\
HierMTN-CRF \citep{wu2020hierarchical}        & 55.72          & 65.26          & 33.91          & 52.34          & -              & -              \\
BERT-FT \citep{kishimoto-etal-2020-adapting}  & 58.48          & 65.26          & -              & 54.32          & -              & -              \\
DS-CP \citep{kurfali2021let}                  & 59.24          & -              & 39.33          & 55.42          & -              & -              \\
BMGF-RoBERTa \citep{liu2020importance}        & 63.39          & 69.06          & 37.95          & 58.13          & \underline{57.26}          & \underline{55.19}          \\
LDSGM \citep{wu2021label}                     & \underline{63.73}          & \underline{71.18}          & \underline{40.49}          & \underline{60.33}          & -              & -              \\
\midrule
PCP w/ RoBERTa-base (PCP-base) & 64.95 & 70.84 & 41.55 & 60.54 & 60.98 & 57.31 \\
PCP w/ RoBERTa-large (PCP-large) & \textbf{67.79} & \textbf{73.80} & \textbf{44.04} & \textbf{61.41} & \textbf{63.36} & \textbf{58.51} \\
\bottomrule
\end{tabular}}
\caption{Experimental results on PDTB 2.0 and CoNLL16. The best results of previous baselines are underlined. Models in the first part of table are based on neural networks and others in the second part (including our method) are based on different PLMs. We have bolded the best performance results in each of the two parts.}
\label{table:overall_comparison}
\end{table*}

\subsection{Baselines}
To validate the effectiveness of our method, We compare our method with previous state-of-the-art methods. First of all, we select some strong baselines based on neural network including NNMA \citep{liu2016recognizing}, ESDP \citep{wang2016two}, MANN \citep{lan2017multi}, PDRR \cite{dai2018improving} and RWP-CNN \citep{varia-etal-2019-discourse}. Their work mainly focused on the top-level senses of PDTB 2.0 and CoNLL16 cross-level senses. Secondly, we compare our method with competitive baselines based on PLMs, such as DER \citep{bai2018deep}, ELMo-C\&E \citep{dai2019regularization}, MTL-MLoss \citep{van2019employing}, HierMTN-CRF \citep{wu2020hierarchical}, BERT-FT \citep{kishimoto-etal-2020-adapting}, DS-CP \citep{kurfali2021let}, BMGF-RoBERTa \citep{liu2020importance} and LDSGM \citep{wu2021label}. These methods achieve impressive performance at the fine-grained second-level senses with the help of large-scale PLMs.

\subsection{Implementation Details}
In this work, we use \emph{RobertaForMaskedLM} as the backbone of our method, in which \emph{RobertaEncoder} is used for obtaining context representation of inputs and \emph{RobertaLMHead} is to acquire each vocabulary token prediction score for <mask> token position. About prompt settings, we select the last template in Table \ref{table:template} and answer set in Table \ref{table:mapping_pdtb} and \ref{table:mapping_conll} to perform our expreiments on PDTB 2.0 and CoNLL16 dataset.

We adopt AdamW optimizer \citep{loshchilov2017decoupled} with the learning rate of $1e^{-5}$ and weigh decay coefficient of $1e^{-4}$ to update the model parameters and set batch size as 4 for training and validation. At the same time, we add label smoothing with a coefficient of 0.05 into the cross-entry loss function to alleviate overfitting. Based on the above settings, we can usually get the best performance results in the first 3 epochs. All our experiments were performed on one RTX 3090. All other parameters are initialized with the default values in PyTorch Lightning\footnote{\url{https://github.com/Lightning-AI/lightning}} and our model are all implemented by transformers\footnote{\url{https://github.com/huggingface/transformers}}. 

\subsection{Experimental Results and Analysis}
We first evaluate our model on the four coarse-grained top-level and 11 fine-grained second-level senses (denoted as PDTB-Top and PDTB-Second) of PDTB 2.0 with Macro-F1 score and accuracy value. Then we conduct 15-class classification on the CoNLL16 dataset and consider accuracy as the main metric, denoted as CoNLL-Test and CoNLL-Blind for the test and blind-test set. 

\begin{table*}[tbp]
\centering
\resizebox{\textwidth}{!}{
\begin{tabular}{c|c|c|c|c}
\toprule
\textbf{Second-level Senses} & \textbf{BMGF-RoBERTa} & \textbf{LDSGM} & \textbf{PCP-large} & \textbf{PCP-large w/o </s>} \\
\hline
Comp.Concession      & 0.0           & 0.0   & 8.00    & \textbf{14.81}   \\
Comp.Contrast        & 59.75         & \underline{63.52} & \textbf{63.88}   & 63.49   \\
\hline
Cont.Cause           & 59.60         & \underline{64.36} & \textbf{65.64}   & 64.90   \\
Cont.Pragmatic cause & 0.0           & 0.0   & 0.0     & 0.0     \\
\hline
Expa.Alternative     & 60.0          & \underline{63.46} & 66.67   & \textbf{70.00}   \\
Expa.Conjunction     & \underline{60.17} & 57.91         & 57.78   & \textbf{58.35}   \\
Expa.Instantiation   & 67.96         & \underline{72.60} & \textbf{74.01}   & 73.39   \\
Expa.List            & 0.0           & \underline{8.98}  & 29.63  & \textbf{37.50}    \\
Expa.Restatement     & 53.83         & \underline{58.06} & \textbf{61.00}   & 55.32   \\
\hline
Temp.Asynchronous    & 56.18         & \underline{56.47} & 57.81   & \textbf{60.00}   \\
Temp.Synchrony       & 0.0           & 0.0   & 0.0     & \textbf{12.50}   \\
\hline
Macro-F1             & 37.95         & \underline{40.49} & 44.04   & \textbf{46.38}   \\
Acc                  & 58.13         & \underline{60.33} & \textbf{61.41}   & 60.64   \\
\bottomrule
\end{tabular}}
\caption{Experimental results on PDTB 2.0 second-level senses. The best results of previous baselines are underlined and the best performance of our method is bolded. We also show changes in the results of our approach (PCP-large and PCP-large without segment token </s>) compared to previous best model.}
\label{table:second_level}
\end{table*}

Table \ref{table:overall_comparison} shows the main experimental results of our method and other baselines. Intuitively, our method achieves the new SOTA performance with substantial improvements for coarse-grained and fine-grained implicit discourse relation recognition, which demonstrates that prompt-based connective prediction can effectively mine specific knowledge about connectives and discourse relations in the large-scale PLMs and improve the model's ability to recognize implicit discourse relations.

To better evaluate the performance of our method on fine-grained implicit discourse relation recognition, we compare it with two previous competitive models at each second-level sense of PDTB 2.0. As shown in Table \ref{table:second_level}, our PCP approach surpasses the previous state-of-the-art model in almost all second-level senses, except for \textit{Expa.Conjunction}. Noticeable improvements were also achieved by our method on two few-shot second-level senses (i.e. \emph{Comp.Concession} and \emph{Expa.List}). 

In addition, in order to further explore our model performance on some few-shot senses, we removed the segment token </s> from the inputs to make it more consistent with the pre-training step of the RoBERTa (corresponding to the forth template in Table \ref{table:template}). The new experimental results are shown in the last part of Table \ref{table:second_level}. The results indicate that our model without segment token breaks the bottleneck of previous work in three few-shot second-level senses (i.e. \emph{Comp.Concession}, \emph{Expa.List} and \emph{Temp.Synchrony}) of PDTB 2.0. We achieved significant improvements in these few-shot senses.

\subsection{Case Study}

For the fine-grained sense \textit{Cont.Pragmatic cause}, our method and previous state-of-the-art model predict incorrectly for all samples. Therefore, we check our model prediction results for all seven samples of this sense in test set, and the final results statistics are summarized in Table \ref{table:case_study}.

We can see that our method tends to predict the connectvies of these samples as high-frequency connectvies, like \textit{because} and \textit{and}, that belong to other discourse relations. So after the answer mapping, the model predicts them as other second-level senses. However, after successively examining these samples, we found that the results predicted by our method seemed to be closer to its actual discourse relation. In other words, the actual discourse relation of these samples that annotated as \textit{Cont.Pragmatic cause} is more similar to the \textit{Cont.Cause} sense. This is probably the reason for merging \textit{Cont.Pragmatic cause} into \textit{Cont.Cause} in CoNLL16 dataset.

\begin{table}[tbp]
\centering
\resizebox{\linewidth}{!}{
\begin{tabular}{c|c|c}
\toprule
\textbf{Mapped Senses}       & \textbf{Predicted Connectvies} & \textbf{Count} \\
\hline
Exp.Restatement              & indeed                & 1     \\
\hline
Exp.Conjunction              & and                   & 2     \\
\hline
\multirow{2}{*}{Cont.Cause} & because               & 3     \\
\cline{2-3}                 & as                    & 1     \\
\bottomrule                                   
\end{tabular}}
\caption{Statistics on the prediction results of our method for the second-level sense \textit{Pragmatic cause}.}
\label{table:case_study}
\end{table}

\section{Discussion}
\subsection{Why not Predict Discourse Relations Directly}
\label{whynot}
Our approach to implicit discourse relation recognition is prompt-based connective prediction. What happens if we use prompt directly to predict implicit discourse relations. To figure out this question, we construct a new template and search new answer set for prompt-based implicit discourse relation prediction (\textbf{PIDRP}). Specifically, the new template is shown below:
\begin{itemize}
    \item Arg1: \textit{Arg1}. Arg2: \textit{Arg2}. The discourse relation between Arg1 and Arg2 is <mask>.
\end{itemize}
and we directly use the top-level senses as their answer sets (i.e. mapping the sense \textit{Comparison} to answer set [comparison]). 

We use the same pre-trained model and data processing on top-level senses of PDTB to compare the performance gap between the two methods. As shown in Table \ref{table:direct_relation}, the PIDRP method 
performs worse than the PCP method on all four top-level senses of the PDTB, especially on the \emph{Temporal} sense. We think that the main reason of poor performance is that connective prediction is closer to the natural language patterns when the model is in pre-training stage than direct implicit discourse relation prediction.

\begin{table}[htbp]
\centering
\begin{tabular}{ccc}
\toprule
\textbf{Top-level Senses} & \textbf{PCP} & \textbf{PIDRP}  \\ 
\midrule
Comparison           & 70.38          & 65.26           \\
Contingency          & 64.18          & 64.02           \\
Expansion            & 80.17          & 79.80           \\ 
Temporal             & 56.41          & 39.13           \\ 
\midrule
Macro-F1             & 67.79          & 62.05           \\
\bottomrule
\end{tabular}
\caption{Performance comparison between PCP and PIDRP on four top-level senses of PDTB 2.0.}
\label{table:direct_relation}
\end{table}

\subsection{Generalization to Explicit Discourse Relation Recognition}
Inspired by the attempt of section \ref{whynot}, we transfer our PCP method to explicit discourse relation recognition. We want to explore how well the prompt-based method predicts discourse relation when given connectives. 

Similarly, we have designed a simple template in line with natural language conventions for the prompt-based explicit discourse relation recognition (\textbf{PEDRR}). The new template is as follows:
\begin{itemize}
    \item Arg1: \textit{Arg1}. Arg2: \textit{Arg2}. The connective between Arg1 and Arg2 is \emph{Connective}. In summary, the discourse relation between Arg1 and Arg2 is <mask>.
\end{itemize}
where the \emph{Connective} represents connectives that appear in the original text but ont in \textit{Arg1} or \textit{Arg2}.

\begin{table}[htbp]
\centering
\resizebox{\linewidth}{!}{
\begin{tabular}{c|c|c}
\toprule
\textbf{Top-level}                    & \textbf{Second-level}    & \textbf{Answer Set}    \\
\hline
\multirow{2}{*}{Comparison}  & Concession      & concession    \\ \cline{2-3}
                             & Contrast        & contrast      \\
\hline
\multirow{2}{*}{Contingency} & Cause           & cause         \\ \cline{2-3}
                             & Pragmatic cause & justification \\
\hline                             
\multirow{5}{*}{Expansion}   & Alternative     & alternative   \\ \cline{2-3}
                             & Conjunction     & conjunction   \\ \cline{2-3}
                             & Instantiation   & instance      \\ \cline{2-3}
                             & List            & list          \\ \cline{2-3}
                             & Restatement     & repetition    \\
\hline                             
\multirow{2}{*}{Temporal}    & Asynchronous    & asynchronous  \\ \cline{2-3}
                             & Synchrony       & simultaneous  \\
\bottomrule                             
\end{tabular}}
\caption{Mapping between explicit discourse relation labels and connectives on PDTB 2.0 dataset.}
\label{table:edrr_mapping}
\end{table}

We are equal to regard the second-level sense as the corresponding answer set (detailed in Table \ref{table:edrr_mapping}). But for some special senses in which their output of RoBERTa tokenizer is not a single token, we find synonyms for them (e.g. \textit{Instantiation}, \textit{Restatement} and \textit{Synchrony}) or use the corresponding subtype as the answer word (e.g. \textit{Pragmatic cause}) to ensure that the result after tokenization is a single token.

As shown in Table \ref{table:pedrr_pdtb}, the variant of our method PEDRR is also able to achieve results close to those of previous SOTA models on the top-level senses of PDTB 2.0, which effectively demonstrates the generalizability of our approach to EDRR.

\begin{table}[htbp]
\centering
\resizebox{\linewidth}{!}{
\begin{tabular}{lcc}
\toprule
\textbf{Model}                        & \textbf{Acc.} & \textbf{F1}      \\
\midrule
(1)Connective Only \citep{pitler-nenkova-2009-using}     & 93.67    & -             \\
(1)+Syntax+Conn-Syn \citep{pitler-nenkova-2009-using}    & 94.15    & -             \\
(2)ELMo-C\&E  \citep{dai2019regularization}              & 95.39    & 94.84         \\
(3)RWP-CNN \citep{varia-etal-2019-discourse}             & \textbf{96.20}    & \textbf{95.48}         \\
\midrule
PEDRR (Ours)                       & 94.78    & 93.59         \\
\bottomrule
\end{tabular}}
\caption{Experimental results of our PEDRR method and other strong baselines on PDTB 2.0 top-level senses for EDRR.}
\label{table:pedrr_pdtb}
\end{table}

\section{Conclusion}
In this paper, we propose a novel prompt-based connective prediction method for coarse-grained and fine-grained implicit discourse relation recognition. Experimental results demonstrate that our method achieves state-of-the-art performance on the PDTB 2.0 dataset and the CoNLL-2016 Shared Task. Furthermore, our proposed method breaks the bottleneck of previous work in the few-shot fine-grained discourse relation of PDTB 2.0. Finally, we experimentally prove that our approach can be transferred from IDRR to EDRR and still performs well for EDRR. We will later explore the applicability of our approach to some Chinese discourse relations datasets for fine-grained DRR.

\section*{Limitations}
In this section, we will point out the limitations of our work, which can be summarized in the following two aspects. 

Firstly, in the step of prompt construction (Section \ref{sec:prompt_construction}), we manually design templates that meet the task goal of connective prediction and follow natural language patterns. But we do not make more attempts at other templates. We think there may be a new template more suitable for this task goal, which could achieve better performance. Furthermore, we have not explored how well template ensemble works. Therefore, this is a crucial research area in our future work.

Secondly, the data statistics in Table \ref{table:pdtb_second} show a category imbalance issue in the second-level senses of PDTB 2.0. The experimental results in Table \ref{table:second_level} also show that our method does not solve this problem well, as it markedly improves the model performance on few-shot classes while there is a slight decrease on "more-shot" classes. In the future, we will explore new methods that can significantly improve the model performance in fine-grained implicit discourse relation recognition on few-shot classes while also improving the model performance on "more-shot" classes.

\section*{Acknowledgements}
We appreciate the support from National Natural Science Foundation of China with the Research Project on Human Behavior in Human-Machine Collaboration (Project Number: 72192824) and the support from Pudong New Area Science \& Technology Development Fund (Project Number: PKX2021-R05).

\bibliography{anthology,custom}
% Prompt Learning
\nocite{liu2021pre,
        schick2020exploiting,schick2020small,chen2022adaprompt,jiang2022promptbert,
        Han2021PTRPT,chen2022knowprompt,cui2021template,chen2021lightner,hu2021knowledgeable}

% Discourse Relation
\nocite{
    xiang2022survey,kim2020implicit,Ji2015OneVI,
    zhou2010effects,lin2009recognizing,pitler-nenkova-2009-using,
    liu2016recognizing,wang2016two,lan2017multi,dai2018improving,varia-etal-2019-discourse,
    bai2018deep,dai2019regularization,
    kishimoto-etal-2020-adapting,kurfali2021let,
    van2019employing,wu2020hierarchical,liu2020importance,wu2021label}

% pre-trained Model and dataset
\nocite{devlin2018bert,liu2019roberta,brown2020language,prasad-etal-2008-penn,xue2016conll}

% downstream tasks
\nocite{xiong2019modeling,mihaylov2019discourse,xu2019discourse}

\appendix

\end{document}